\documentclass[runningheads]{llncs}

\usepackage{eccv}

\usepackage[dvipsnames]{xcolor}

\usepackage{tabularx,multirow,booktabs}
\usepackage{newunicodechar}
\usepackage{graphicx}
\usepackage{tikz}
\usepackage{pgfplots}
\usepackage{array}
\usepackage{pifont}
\usepackage{comment}
\usepackage{bm}

\newcommand{\pname}{\textit{Disagreement Reconciliation}\xspace}
\newcommand{\aname}{\textit{Look Around}\xspace}
\newcommand{\baname}{\textbf{Look Around}\xspace}

\let\titleold\title

\renewcommand{\title}[1]{\titleold{#1}\newcommand{\thetitle}{#1}}

\def\maketitlesupplementary
   {
   \onecolumn
   \begin{center}
   \Large
    \textbf{\thetitle}\\
    \vspace{0.5em}Supplementary Material \\
    \vspace{1.0em}
   \end{center}
   }

\usepackage{eccvabbrv}

\usepackage{graphicx}
\usepackage{booktabs}

\usepackage[accsupp]{axessibility}  %

\usepackage{hyperref}

\usepackage{orcidlink}

\title{\textit{Look Around} {\Huge{\includegraphics[]{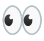}}} and Learn {\Huge{\includegraphics[]{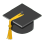}}}: Self-Training Object Detection by Exploration}

\titlerunning{Look Around and Learn}

\author{Gianluca Scarpellini\inst{1,2}\orcidlink{0000-0002-3468-8902} \and
Stefano Rosa\inst{1}\orcidlink{0000-0001-6458-6344} \and
Pietro Morerio\inst{1}\orcidlink{0000-0001-5259-1496} \and
Lorenzo Natale\inst{1}\orcidlink{0000-0002-8777-5233} \and
Alessio Del Bue\inst{1}\orcidlink{0000-0002-2262-4872}
}

\authorrunning{G.~Scarpellini et al.}

\institute{Istituto Italiano di Tecnologia, Genoa, Italy\\ \email{\{name.surname\}@iit.it} \and
University of Genoa, Genoa, Italy}

\begin{document}
\maketitle

\begin{abstract}
When an object detector is deployed in a novel setting it often experiences a drop in performance. This paper studies how an embodied agent can automatically fine-tune a pre-existing object detector while exploring and acquiring images in a new environment without relying on human intervention, i.e., a fully self-supervised approach.
In our setting, an agent initially learns to explore the environment using a pre-trained off-the-shelf detector to locate objects and associate pseudo-labels.
By assuming that pseudo-labels for the same object must be consistent across different views, we learn the exploration policy ``\textbf{Look Around}'' to mine hard samples, and we devise a novel mechanism called ``\textbf{Disagreement Reconciliation}'' for producing refined pseudo-labels from the consensus among observations. 
We implement a unified benchmark of the current state-of-the-art and compare our approach with pre-existing exploration policies and perception mechanisms. Our method is shown to outperform existing approaches, improving the object detector by $6.2\%$ in a simulated scenario, a $3.59\%$ advancement over other state-of-the-art methods, and by $9.97\%$ in the real robotic test without relying on ground-truth.

Code for the proposed approach and baselines are available at \url{https://iit-pavis.github.io/Look_Around_And_Learn/}.

\end{abstract}
    
\section{Introduction}
\label{sec:introduction}
\begin{figure}[t]
\begin{center}
\includegraphics[width=0.7\linewidth]{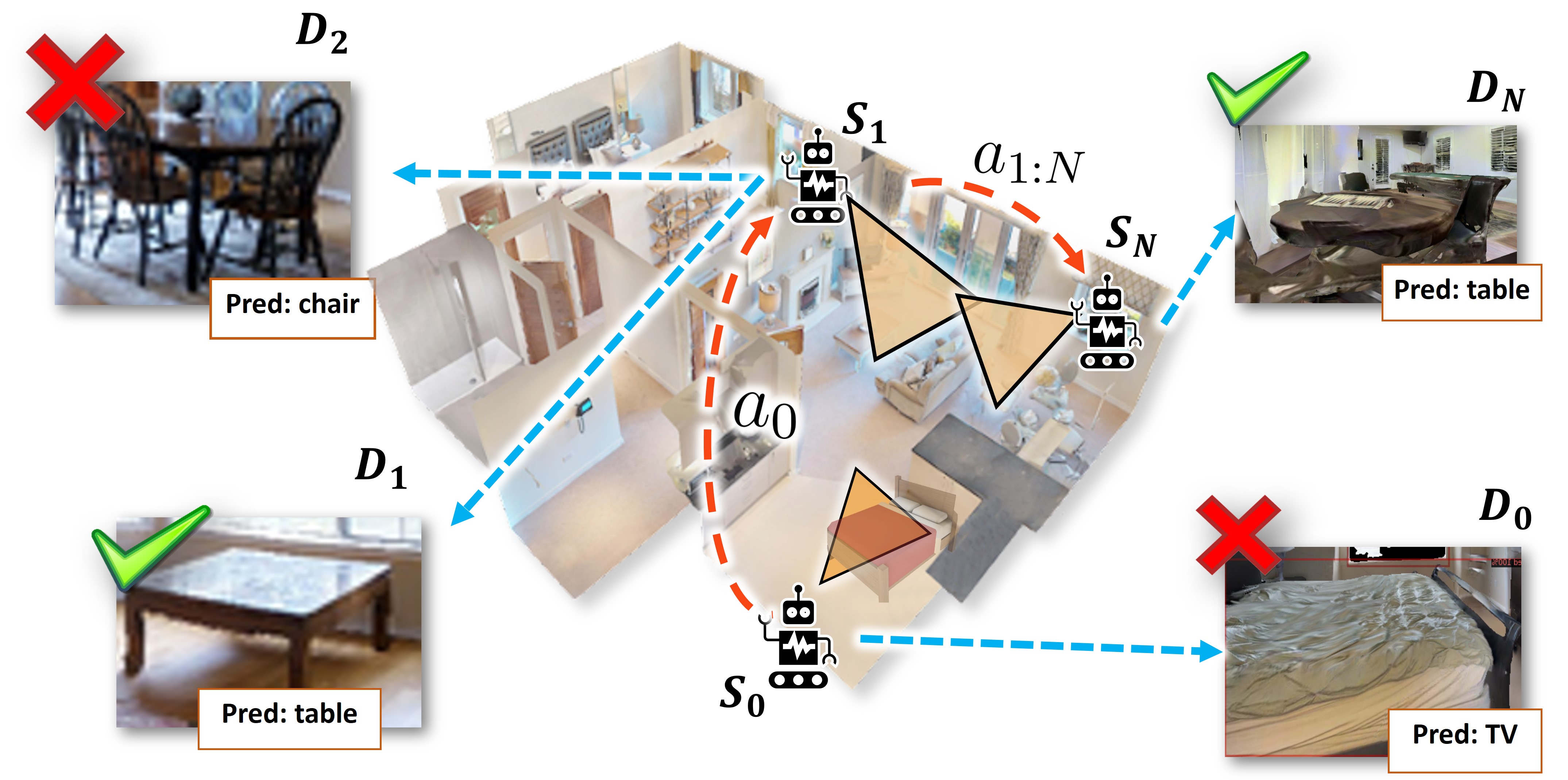}
  \caption{We equip an agent with an off-the-shelf object detector. The agent explores a new environment and collects a set of noisy detections $D_{0:N}$ following a trajectory $S_{0:N}$ resulting from following intermediate goals $a_{0:N}$ predicted by the exploration policy $\pi$. Such detections are then used for finetuning the detector.}
   \label{fig:fig1}
\end{center}
\vspace{-10pt}
\end{figure}

When newborns interact with novel objects, they are naturally inclined to actively explore their hidden facets by manipulating or moving around to have views from a new angle, thus unveiling blindsides. Yet, a child is aware that, no matter the viewpoint, the object instance is always the same. Indeed, findings in social science suggest that humans and animals develop perception and learn by actively exploring an environment \cite{gibson1950perception} in a so-called action-perception loop \cite{action-perception}, with very little or no external supervision. 
Inspired by human perception and by recent advances in active learning \cite{yang2019embodied,chaplot2020semantic,chaplot2021seal}, in this work, we posit that the process of unsupervised tuning of object detectors can indeed benefit from an active exploration of the environment (Figure \ref{fig:fig1}). 

We suggest that, by \textit{looking around} the environment and seeing the same object instances from different views, the agent can \textit{learn} and improve its object detection skills. 
For this reason, we introduce a learnable policy that guides an embodied agent to improve object detection without external supervision by directly rewarding the acquisition of uncertain samples. Our proposal extends previous literature on environment exploration, which mostly leverages reinforcement learning as a way of maximizing the total area explored in the process \cite{chaplot2020learning}, the novelty of the observation \cite{ramakrishnan2020exploration}, or success in reaching a predefined goal \cite{zhu2017target,chen2021batch}. 

The proposed pipeline includes learning both \textit{action} (dubbed \aname) and \textit{perception} (dubbed \pname) stages in a completely self-supervised fashion, and it assumes an agent exploring an unknown environment using an off-the-shelf object detector.

The action phase accumulates predictions in a 3D map, with policy rewards based on maximizing the detector's prediction disagreement. The perception phase then refines these predictions as pseudo-labels for fine-tuning the detector, focusing on consistency across views, contrastive learning, and aggregating multiple views. A perception module (object detector) driven by active exploration somehow contrasts with the most classic paradigm of computer vision, where models are usually trained with static datasets and do not fully benefit from the realistic continuity of the 3D world. Our perception module adopts its predictions as \textbf{pseudo-labels} to improve performance without external supervision.

We test the improved object detector on a dataset we collected from unseen scenes. Our experiments in the simulated environment show that \aname with \pname successfully improves the object-detector by 6.3\% mAP without relying on external supervision. Our experiments also show that finding hard examples to feed the detector leads to more informative pseudo-labels, resulting in more accurate object detectors. 

We further validate our proposed approach by deploying \aname on a wheeled humanoid robotic platform. We deploy the learned policy on the robot without further adaptation and fine-tune the robot's object-detector via \pname. Our findings show that fine-tuning the object-detector via \aname with \pname improves the object-detector in a real scenario by 9.97\% mAP without any ground truth or human interventions. We are the first to show that self-training by exploration works in a real scenario.

In addition to these contributions, a significant aspect of this work is the development of a benchmark for evaluating self-training with active perception. Accompanied by publicly available code and dataset, our framework provides a comprehensive benchmark for assessing various action policies, disagreement metrics, and perception modules. We implement state-of-the-art policies \cite{chaplot2021seal,jing2023learning,chaplot2020learning,chaplot2020object,yamauchi1998frontier} and perception methods \cite{chaplot2021seal,xie2020selftraining}, and we consistently evaluate those method on environments \cite{xiazamirhe2018gibsonenv} in the Habitat simulator \cite{szot2021habitat}. 

To summarize, the main contributions of this work are:
\begin{itemize}

\item A fully self-supervised method for fine-tuning object detection for embodied agents, composed of a policy that mines inconsistent detections and a novel probabilistic disagreement measure to compute consistent pseudo-labels.

\item A publicly available framework for benchmarking across different action policies, disagreement metrics, and perception modules. Our extensive experiments demonstrate the ability of \aname\ and \pname\ to achieve state-of-the-art results under various settings.

\item Validation of the proposed method through real-world testing on a robotic platform, improving the object detector in practical settings.
\end{itemize}

\section{Related work}
\label{sec:relatedwork}

\paragraph{\textbf{Pseudo-labeling in object detection.}}%
Pseudo-labeling is often used in self-supervised learning, which is defined as adopting predictions from a neural network as labels for unlabeled data \cite{lee2013pseudo}. However, simply using neural network outputs suffers from confirmation bias and high-class imbalance \cite{arazo2019}. Similar behaviors are commonly observed in object detectors, where pseudo-labels are hard to exploit due to foreground-background imbalance \cite{liu2021unbiased}.
Inspired by previous works on consistency training \cite{xie2020self,xie2019unsupervised}, we propose to solve for the confirmation bias by correcting the pseudo-labels through a consensus mechanism. 

\paragraph{\textbf{Self-training.}}
\label{lit:st}
Our method shares similarities with the self-training paradigm, where a model is trained jointly on labeled and unlabeled data \cite{lee2013pseudo}. A pre-trained model is used to generate pseudo-labels for unlabeled data. Recent literature suggests generating pseudo-labels with noise-free models and data while training with strong data augmentation and noise injection (e.g., dropout) \cite{arazo2019,xie2020selftraining,chen2020big}. 
Chaplot \etal \cite{chaplot2021seal} proposed an approach for collecting consistent pseudo-labels in an environment. A semantic map of the environment is built and re-projected onto each frame to get consistent pseudo-labels across views. 
We introduce a novel mechanism to leverage different points of view of the same object as a data-augmentation strategy. We adopt a contrastive loss \cite{facenet2015} to impose that features for an object should be alike across views, while different objects should be distinguishable in the feature space. In this sense, we interpret our approach as a generalization of the cropping strategy for view generation, which is quite popular in self-supervised learning \cite{chen2020simple}.

\paragraph{\textbf{Object detection for active visual exploration.}}
\label{lit:active}
Our proposal can be linked with \textit{active perception} for \textit{visual exploration} of the environment. 
Active perception can represent a broad spectrum of concepts \cite{bajcsy2018revisiting}. Still, in the context of robotics, it describes the problem of actively moving in an environment to collect useful examples and subsequently improve the performance on a given task. 
Chaplot et al. \cite{chaplot2020semantic} proposed to learn exploration policies for online data collection. There are two main differences between our proposal and theirs. %
First, \cite{chaplot2020semantic} divides the 3D scene into 2D cells (top-view) and counts the number of predictions for each cell. It assumes that, at most, one object can fit in each cell, while our method computes the disagreement score for each distinct object in 3D space before building the disagreement map. Second, they rely on ground-truth annotation for training the object-detector, while our method is completely self-supervised.
Chaplot \etal \cite{chaplot2021seal} train an exploration policy based on the number of confident object predictions, but they do not accumulate predictions and reward the number of 3D cells containing conflicting predictions.
Recently, Jing \etal \cite{jing2023learning} proposed to build a 3D semantic map of detections and reward the agent based on the Kullback-Leibler divergence between predictions from current observations and the corresponding class distribution on the semantic map.
Contrary to these approaches, we adopt inconsistency of the detections as a viable training signal for both policy and object-detector.
Fang \etal \cite{fang2020move} collect multiple views for one object to improve the object-detector, but cannot be adopted for exploring an entire scene. Other methods assume access to the test scene at inference for improving the object detector \cite{kotar2022interactron}.
Min \etal \cite{self-supervised-object-goal} propose fine-tuning a semantic segmentation model by collecting samples via frontier exploration. 

We propose to explore a set of environments via \aname to train a policy and fine-tune the object detector, and we test the fine-tuning on a test set of images from unseen environments.

\section{Methodology}
\label{sec:methodology}
\begin{figure*}[tb]

\begin{center}
    \includegraphics[width=0.99\linewidth]{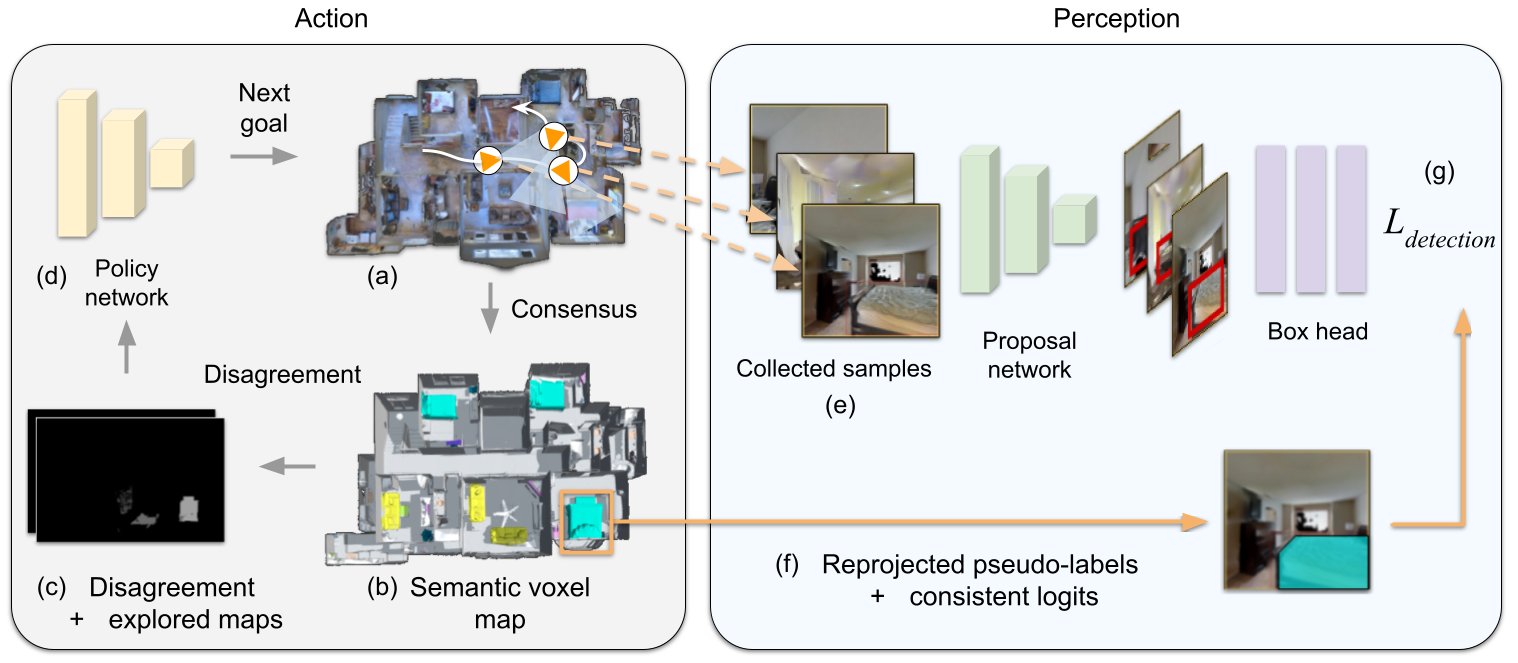}
   \caption{Our approach consists of an action and a perception phase. \textbf{\textit{Action:}} \textbf{(a)} our policy predicts a long-term goal for the agent. \textbf{(b)} During the exploration, the agent builds a semantically consistent voxel map of the environment by projecting the predictions of its object detection into a tri-dimensional space. \textbf{(c)} We project the map onto top-down view and compute a disagreement map by assigning a disagreement score value to each cell. \textbf{(d)} The disagreement map is the input of our policy network. \textbf{\textit{Perception:}} \textbf{(e)} we collect samples by exploring an environment via the learned policy and \textbf{(f)} project the semantic voxel map onto each observation to build the pseudo-labels for the self-training scheme. \textbf{(g)} Finally, we fine-tune the object detector by relying only on the pseudo-labels. 
   }
   \label{fig:fig2}
\vspace{-1cm}
\end{center}
\end{figure*}

We equip real and simulated agents with an object detector, an RGB-D sensor, and a position sensor. The agent \textit{looks around} and collects samples from a set of simulated environments \cite{xiazamirhe2018gibsonenv} in the Habitat simulator \cite{szot2021habitat} or in the real world. A policy sets the long-term goals for the agent, which moves toward the goal via a path planner. %
We aim to fine-tune the agent's object detector by relying only on the autonomously generated samples, i.e., without ground-truth annotations. 
To be comparable with previous literature \cite{chaplot2020semantic,chaplot2021seal}, we adopt a 2-stage object-detector \cite{he2018mask} with a Resnet50 backbone \cite{resnet} as implemented in \cite{wu2019detectron2} and pre-trained on the COCO dataset \cite{lin2014microsoft}. We acknowledge that 3D object detectors and point-cloud-based methods \cite{ahmadyan2021objectron, mohammadi2021pointview} would directly provide the 3D point cloud for the predictions; nonetheless, we adopt a 2D object detector in the exploration loop, as our proposal exploits the uncertainty of the object detector itself to improve its performance after the fine-tuning stage. 
We now formalize and provide further details on the two stages of our methodology: \aname~(action stage) and \pname~(perception stage) (Figure \ref{fig:fig2}).

\subsection{Action: Look Around}
\label{sec:method_policy}
The action stage is formalized as a Markovian Decision Process, and the objective becomes learning a policy $\pi$ that, given the current observation $S_i$, predicts the next action to take $a_i$. Each state $S_i$ initially contains the input RGB-D image $x_i$, the map of the explored environment $M_i$ and object detector's predictions $D_i$ --- a set of $n$ bounding-boxes $b_1 \dots b_n$, instance segmentation masks $m_1 \dots m_n$, predicted classes $y_i$ and normalized logits $\lambda_1 \dots \lambda_n$. 

During the action phase, we train a policy $\pi$ parametrized as a neural network to predict long-term goals.
We first compute the disagreement score of the object detector, which measures the entropy of the detections for all the detected instances.
We project the disagreement score in a 3D semantic map (Figure \ref{fig:fig2}{\color{red}b}) and reproject the 3D map on a top-down 2D disagreement map (Figure \ref{fig:fig2}{\color{red}c}).
Next, we feed the neural network $\pi$ with the disagreement map to predict the next long-term goal (Figure \ref{fig:fig2}{\color{red}d}). A planner produces a set of optimal sub-goals to reach the final goal. The agent takes actions toward each intermediate goal for $N_\text{replanning}$ steps before predicting a new long-term goal (Figure \ref{fig:fig2}{\color{red}a}). We train $\pi$ to maximize the total disagreement score of the object detector via reinforcement learning.
We now provide an in-depth analysis of these components.

\paragraph{\textbf{Semantic voxel map.}}

We propose to leverage tri-dimensional projection to build a semantic voxel map from noisy predictions (Figure \ref{fig:fig2}{\color{red}b}).

We exploit the depth and position of the sensors of the agent to build a semantic point cloud by projecting the agent's detector predictions $D_i=(b_i, m_i, y_i, \lambda_i)$ (respectively, instance segmentation masks, classes, and logits from the 2D detector) to 3D coordinates. First, we voxelize the resulting point cloud with $\mathtt{voxel\_size}=0.05m$. In the resulting voxel map, each non-empty voxel maintains the set of all logit vectors.  We then compute the hard label $\overline y$ of a voxel as the class with the maximum score among all the predictions associated with that particular voxel (Figure~\ref{fig:fig1}). We highlight, however, that this voxelization process is \textit{lossy}, as we compute $\overline y$ via a \textit{max} operator.
We aggregate connected voxels with equal classes $\overline y$ via a 26 connected-components algorithm \cite{cc3d}. Each aggregated component is a distinct object instance; therefore, it is uniquely assigned an identifier $u$ with a unique class (the same as the aggregated voxels), and a set of vector logits, and bounding boxes, that will be used to compute the disagreement map and the pseudo-labels in the Disagreement Reconciliation step.

\paragraph{\textbf{Disagreement map.}}
We propose to measure the disagreement of the object detector as viable information for the policy. As we gather the object detections into a semantic voxel map, we preserve all the object detector's predictions --- normalized logits and class predictions --- or each unique 3D instance object $u$. Next, we compute the disagreement score as the entropy of the average logits assigned to an object, such as
\begin{equation}
s_\text{disag-entropy}(u) = \mathbb{E}[-\log p(u)],
\label{eq:entropy}
\end{equation}
where $p(u)$ is the average normalized logits for the unique instance object $u$. This formulation is inspired by the literature on epistemic uncertainty estimation \cite{gal2016dropout,abdar2021review}.
At each time step $i$, we reproject the semantic voxel map onto a 2D \emph{disagreement map} $H_i$ as a top-down map of the environment and add it to the agent's state (Figure \ref{fig:fig2}{\color{red}c}). Each cell $(i, j)$ of the map contains the sum of the disagreement scores $s_\text{disag-entropy}$ of the objects mapped in top-down position $(i, j)$.

\paragraph{\textbf{{Policy network.}}}
To exploit the embodied nature of the agent, a suitable movement policy must be devised for collecting informative new samples.
We parameterize the policy $\pi$ with a neural network.
The policy takes as input $I_i$ as:
\begin{equation}
I_i = \{ S^{map}_i \in [0,1]^{2 \times K \times K}, e_i \},
\end{equation}
where the first channel of $S^{map}_i$ contains the current disagreement map $H_i$ with canonical dimension $K\times K$; the second channel contains the map of the explored environment $M_i$ with the agent's position superimposed (also resized to dimensions $K \times K$); finally $e_i$ contains the global orientation of the agent with respect to the map (Figure \ref{fig:fig2}{\color{red}d}). 

Based on the agent's state, the policy predicts an action $a_i = (x,y)$ that represents a new long-term goal position in the map for the agent to explore.
The goal is then fed to a graph-based path planner ($A^*$) in order to compute a feasible path to the goal (Figure \ref{fig:fig2}{\color{red}a}).
In particular, we first abstract the map to a visibility graph, where each node of the graph is a reachable position in the map, and each edge connects two nodes if they are reachable in a straight line without collision with obstacles. We then adopt the closest reachable node to the policy prediction as the goal position.
Graph search is finally used to extract the shortest node sequence that connects the agent to the goal.
After $N_\text{replanning}$ steps, the policy network predicts a new long-term goal.

\subsection{Perception: Disagreement Reconciliation} 
\label{sec:perception}
During the perception phase, we adopt the policy $\pi$ to collect a sequence of states $S_{0:N}$ for each environment (Figure~\ref{fig:fig2}{\color{red}e}).
We propose a novel mechanism for producing consistent pseudo-labels across different views of the same environment while injecting the original inconsistencies as a soft target to help training further. Finally, we leverage the rich information in our semantic point cloud to fine-tune the object detector. 
Pseudo-labels collected by an agent while navigating in a new environment might be very noisy, and instances may be missed altogether if the object is seen from an unusual point of view. These inconsistencies have to be dealt with during self-training.

During the exploration phase, we build the semantic voxel map as in the action stage. As previously discussed, this process drops important information about the inconsistencies between the predictions. To overcome this issue, we additionally compute for each object instance $u$ an aggregated softmax vector $\overline \lambda_{u}$ as:
\begin{equation}
\label{eq:clean}
\overline \lambda_{u} = \frac{1}{| \mathcal{Q}(u)|}\sum_{\lambda \in Q(u)} \text{softmax}( \lambda ),
\end{equation}
where $\mathcal{Q}(u)$ is the set of predicted vector logits $\{\lambda_i \}$ for the object instance $u$ (Figure \ref{fig:fig2}{\color{red}f}). We further discuss the aggregated softmax vector in the following Sections, where we underline the importance of this soft target for the fine-tuning of the detector.

Any 3D object instance $u$ that appears in the semantic voxel map is projected onto each $x_i$ using the intrinsic and extrinsic camera matrix, resulting in a per-instance identifier mask $\overline m_i$ and aggregated softmax vector $\overline \lambda_{u_i}$ for the projected object instance $u_i$ juxtaposed onto $x_i$. We compute bounding box $\overline b_i$ as the minimum rectangle containing the mask $\overline m_i$. Each pseudo-label contains the object identifier $u_i$, its object's class  $\overline y_{u_i}$, aggregated softmax vector $\overline \lambda_{u_i}$, consistent bounding boxes $\overline b_i$, and consistent masks $\overline m_i$. 
The projection phase solves both issues of the agent's predictions: \textit{(i)} as the voxel map is self-consistent, it follows that pseudo-labels for an object instance are consistent across different views \textit{(ii)} the consistency also applies for frames where no detections were obtained at first.  
Our training set $\mathcal{D}$ consists of the resulting set of observations and pseudo-annotations $\{x_i, (u_i, \overline \lambda_{u_i}, \overline m_i, \overline b_i, \overline y_{u_i})\}_i$. 

\paragraph{\textbf{{Self-training phase.}}}
We propose a fine-tuning strategy for the object detector's head that leverages the consistent pseudo-labels described above (Figure \ref{fig:fig2}{\color{red}g}). 
Given the training set $\mathcal{D}$, we first compute mask, bounding-boxes, and classes \textit{losses} $\mathcal L_\text{head}$ on pseudo-labels $\overline D_0, \dots, \overline D_N$ as in \cite{he2018mask}. To fully exploit the information in our semantic voxel map, we propose two additional steps in the self-training strategy: instance-matching and soft distillation. 
\label{sec:instance}
\begin{figure}[t]
\centering
\includegraphics[width=0.7\linewidth]{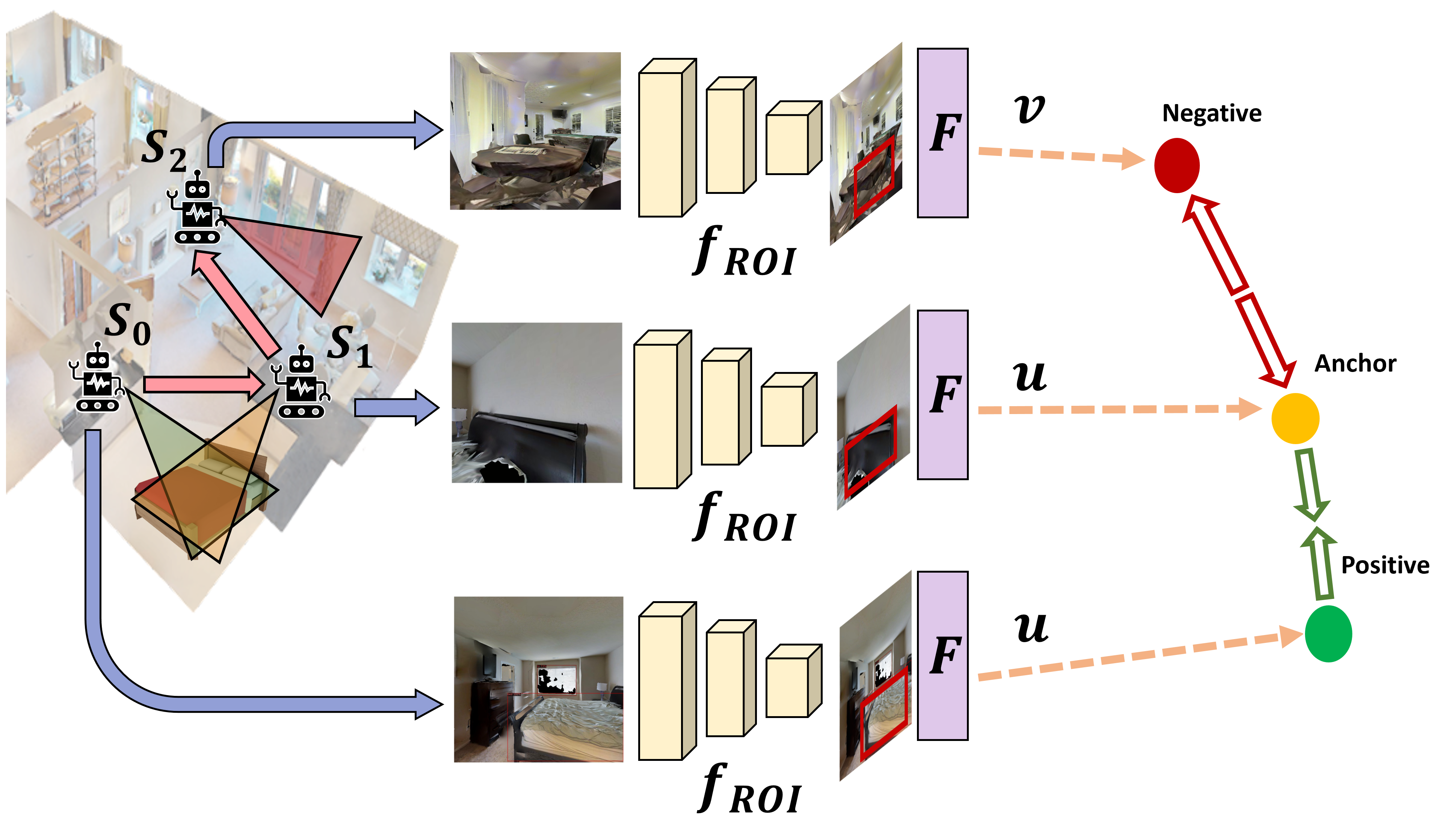}

  \caption{Our policy explores the environment by maximizing the disagreements between predictions for the same object. The instance-matching loss exploits this behavior. It enforces feature vectors belonging to the same object ($\text{u}$ in the Figure) to be close in the feature space while moving away feature vectors of different objects ($\text{u}$ and $\text{v}$).}
  \label{fig:triplet}
\vspace{-.5cm}
\end{figure}

\paragraph{\textbf{Instance-matching.}}
The first component of our strategy is a contrastive triplet loss between detections. Intuitively, a contrastive loss helps $f_\text{box-head}$ construct an embedding space where feature vectors of the same objects are closer than feature vectors of different objects (see Figure~\ref{fig:triplet}).

Given a batch of observations $\{x_j\}_{j=1,\dots,B}$ for batch-size $B$ and corresponding pseudo-labels, we predict the regions of interest $\text{ROIs}_{1:N}$ through the first-stage model $f_\text{ROI}$ of the object detector. Next, we provide the ROIs to the model's head $f_\text{box-head}$ to predict a set of feature vectors $F_{1:N}$, where $F_j \in \mathbb{R}^{k\times1024}$. 
We associate each region of interest $\text{ROIs}_{j}$ with its corresponding pseudo-labels $(u, \overline \lambda_{u}, \overline m, \overline b, \overline y)$. Next, leveraging the object identifiers $u$, we compute positive and negative relationships in $\text{ROIs}_{1:N}$. Two ROIs have a positive relationship if they correspond to the same object $u$ and negative if they correspond to different instances $u$ and $v$. We compute $\mathcal L_\text{im}$ by taking the distance between positive and negative examples as:
\begin{align}
\mathcal L_\text{im}(A, P, N) = \max(d_{AP} - d_{AN} + \epsilon, 0),
\end{align}
where $(A, P, N)$ are feature vectors in $F$ and correspond to anchor, positive, and negative feature vectors respectively, $f_\text{fp}$ is a linear feature-projector, $d_{AP}=||f_\text{fp}(A)-f_\text{fp}(P)||_2$, $d_{AN}=||f_\text{fp}(A)-f_\text{fp}(N)||_2$, and $\epsilon~\in~[0, 1]$ is a safe margin.

\paragraph{\textbf{Aggregated softmax vector as soft pseudo-labels.}}
As discussed in Section \ref{sec:perception}, we expect the model to benefit from an additional loss that guides the prediction towards a smooth average of the logits provided by the off-the-shelf detector from different views. The knowledge distillation loss--our second component--introduces the aggregated softmax vector as a form of soft distillation. We compute $\mathcal L_\text{distil}$ as the cross-entropy between the predicted logits $\lambda_i$ and the aggregated softmax vector $\overline \lambda_{u_i}$.

\paragraph{\textbf{Training.}} To summarize, we propose the combined loss $\mathcal L_\text{detection}= \mathcal L_\text{im} + \alpha \mathcal L_\text{distil} + \mathcal L_\text{head},
$
by merging three different losses: \textit{(i)} a triplet loss based on different views of the same object $\mathcal L_\text{im}$, \textit{(ii)} a soft distillation loss $\mathcal L_\text{distil}$ for leveraging the aggregated softmax vector $\overline \lambda$, and \textit{(iii)} mask, bounding-boxes, and classes losses $\mathcal L_\text{head}$ \cite{ren2016faster} on pseudo-labels ($\overline m$, $\overline y$,$\overline b$).
We proceed to fine-tune the model by jointly minimizing $\mathcal{L}_\text{detection}$, and the standard region proposal loss $ \mathcal{L}_\text{rpn}$ introduced in \cite{ren2016faster}.

\section{Experiments}
\label{sec:experiments}
We devise a set of experiments to test policies and fine-tuning techniques separately. 
For each policy described in Section \ref{sec:policies}, we let an agent move in the environment and collect 7,500 frames from the Gibson dataset \cite{xiazamirhe2018gibsonenv}. We opt for MaskRCNN \cite{he2018mask} as the object detector for comparison fairness with previous approaches \cite{chaplot2021seal,chaplot2020semantic,jing2023learning}, although our proposed approach is model-agnostic.
We report additional training detail in the Supplementary material.

\paragraph{\textbf{{Evaluation.}}} To evaluate our approach, we sample 4,000 testing samples as in \cite{chaplot2021seal}. We adopt the mean Average Precision (mAP) with IOU threshold on bounding-boxes at 0.5 as evaluation score \cite{lin2014microsoft}. 
We stress that no semantic annotations are required during training. We introduce the baseline exploration and perception policies in Section \ref{sec:policies}. Section \ref{sec:results} compares \aname\ with \pname\ with the baselines and supervised labels. For the latter, we use the annotations provided by Habitat and originally proposed in Armani \etal \cite{armeni20193d}. We ablate the main components of our approach in Section \ref{sec:ablations}, and we test \aname\ with \pname\ on a real robotic platform in Section \ref{sec:robot}. %

\subsection{Action and perception methods}
\label{sec:policies}
We compare our \textit{action module} with the following classic and learned policies: 
\textbf{Random goals} -- The agent chooses a feasible random goal in the map and moves to the selected point via path planner. \\ 
\textbf{Frontier exploration} \cite{yamauchi1998frontier} -- Frontier implements a simplified version of classical frontier-based exploration \cite{yamauchi1998frontier}. The agent keeps an internal representation of the explored map, computes goal points of interest at the frontiers of the explored map, selects the next goal greedily, and moves towards it via path planner. \\
\textbf{NeuralSLAM} \cite{chaplot2020learning} -- RL policy based on long-term and short-term goal predictions for maximizing map coverage. A global policy predicts the next long-term goal, while a local policy predicts a sequence of short-term steps. The reward is the percentage of exploration of the environment. \\
\textbf{Semantic Curiosity} \cite{chaplot2020semantic} -- RL exploration policy that predicts local steps to maximize inconsistency between detections projected onto the ground plane. \\
\textbf{SEAL} \cite{chaplot2021seal} -- RL policy for maximizing the number of confident predictions of the detector. The agent builds a voxel map of the environment by leveraging the sensor information. A global policy processes the current voxel map and position information and predicts the next long-term goal, while a local path planner guides the agent toward the goal. The reward is the number of voxels assigned to any possible class with a score above $0.9$. \\
\textbf{Informative Trajectories} \cite{jing2023learning} -- RL policy trained on maximizing the KL divergence between current observations and a 3D semantic voxel map of accumulated predictions as well as to explore objects for which the detector is very uncertain between two classes.

We implement \cite{chaplot2020semantic}, \cite{chaplot2021seal}, and \cite{jing2023learning} to the best of our efforts since no public implementation is available. %
We also compare our \textit{perception module} with two baselines:\\
\textbf{Self-training} \cite{yalniz2019billion} -- The perception phase uses the prediction of the off-the-shelf object detector as pseudo-labels for training without any further processing. Self-training often requires millions of images to provide strong benefits. \\
\textbf{SEAL perception} \cite{chaplot2021seal} -- SEAL aggregates the predictions of the off-the-shelf object-detector into a semantic voxel map by assigning to each voxel the class with the maximum score among the predicted classes. Then, the voxel map is re-projected onto each observation. Finally, the detector is fine-tuned by minimizing the MaskRCNN losses \cite{he2018mask}.\\

\begin{table}[tb] 
\caption{We compare mAP of our action and perception phases with the baselines. Our action-perception loop outperforms the baselines with a mAP@50 of 46.60. As a reference, off-the-shelf MaskRCNN reaches mAP@50 40.33.}
\centering

\begin{tabularx}{\linewidth}{X|cccc}
\hline
\hline
\emph{Policy} & Self-training~\cite{yalniz2019billion} & SEAL perc.~\cite{chaplot2021seal} & \textbf{\textit{Disag. reconc.}} &\textit{GT}\\
\hline\hline
Random & 39.67 &41.19& 41.88& \textit{47.20}\\ 
Frontier Exploration \cite{yamauchi1998frontier} &  40.18 & 41.98 & 43.09 & \textit{45.06}  \\  
NeuralSLAM \cite{chaplot2020learning} & 39.98 & 39.56&40.32& \textit{44.86}  \\  
SemCur \cite{chaplot2020semantic}& 40.23 & 41.06 & 41.37 & \textit{44.67}\\
SEAL \cite{chaplot2021seal} & 39.33 & 43.01 & 42.38& \textit{44.57}\\ 
Inform \cite{jing2023learning} & \textbf{40.25 }
& 44.15 & 43.70 & \textit{45.49}\\
\hline
\textbf{\aname} & 38.66  & \textbf{45.90}&\textbf{46.60}&\textbf{\textit{48.01}}\\ 
\end{tabularx}

\label{tab:results_policies}
\vspace{-.7cm}
\end{table}

\subsection{Results}    
\label{sec:results}
We evaluate our method against different policies and perception methods, and Table \ref{tab:results_policies} reports our results.
As a reference, we assess that the off-the-shelf MaskRCNN \cite{he2018mask} reaches mAP@50 $40.33$ on our test set. 

Our perception method outperforms self-training \cite{yalniz2019billion} and SEAL's perception \cite{chaplot2021seal} in combination with random, frontier, Semantic Curiosity, and NeuralSLAM baselines. The improvement is evident when combining our self-training mechanism and our policy: in this scenario, our pipeline \aname~+ \pname~reaches the overall best result of mAP@50 $46.60$, which is $6.2\%$ over off-the-shelf performance. Even when using the random goals baseline, our perception module exceeds self-training by $2.21\%$. 

Compared to exploration policies, both learned \cite{chaplot2020learning} and classical \cite{yamauchi1998frontier} ones, our perception module outperforms SEAL perception. In particular, we notice that a classical frontier exploration baseline outperforms NeuralSLAM in our self-learning task. %
Intuitively, if the variance of the observations is low, the semantic voxel map reflects the labels predicted by the off-the-shelf detector. 
We believe NeuralSLAM exploration collects fewer observations for each object instance. We report similar results for all the perception methods when NeuralSLAM is adopted. 
On the other hand, a frontier exploration baseline combined with our perception module achieves an AP of $43.09$. Our intuition is that, due to the greedy exploration policy, the agent moves through the same areas multiple times, collecting more views of the same object that are then cleaned by the consensus approach. SEAL perception slightly outperforms our perception when applied in conjunction with SEAL policy and Informative Trajectories\cite{jing2023learning}. SEAL agents find views of an object with the highest prediction score, limiting the number of views per object. %
We underline that our action-perception loop outperforms SEAL's one by $3.59\%$ and Informative Trajectories by $2.45\%$. 

\begin{table}[tb]
\caption{Impact of $\mathcal{L}_{im}$ and $\mathcal{L}_{distil}$ on mAP. For $\mathcal{L}_{im}$, the improvement is significant if applied in conjunction with our policy, confirming the validity of our exploration objective. For $\mathcal{L}_{distil}$, improvements are more significant if the agent collects multiple and diverse predictions, as for our policy.}
\centering
\setlength{\tabcolsep}{8pt} %
\begin{tabularx}{\linewidth}{X|cc|cccc}
\hline
\hline
    \multirow{2}{*}{Policy} &\multicolumn{2}{c}{$\mathcal{L}_{im}$ } &\multicolumn{4}{c}{$\alpha \mathcal{L}_{distil}$ }\\
 & \ding{55} & \ding{51} & \ding{55}  & \textbf{0.1} & \textbf{0.7} & \textbf{1.0}\\
 \hline
 \hline
Random & 41.58& 41.88   & 41.01&41.74& 41.88 &41.32\\
Frontier Exploration\cite{yamauchi1998frontier}&42.32&  43.09 &42.23&42.29&  43.09&43.08\\
NeuralSLAM \cite{chaplot2020learning} &40.05&  40.32  &38.70&39.65&  40.32&40.26\\
Semantic Curiosity\cite{chaplot2020semantic}&42.01&41.37&43.16&42.60&41.37&41.25\\
SEAL\cite{chaplot2021seal}&42.35&  42.38&42.72& 42.28& 42.38 &41.16\\
Inform\cite{jing2023learning} & 43.30 & 43.70 & 44.00 & 44.44 & 43.70 & 42.54\\
\hline
\textbf{Ours} &44.19&  46.60 &41.09&44.55&  46.60&41.45\\
\end{tabularx}
\label{tab:abl}
\vspace{-.4cm}
\end{table}

When our policy is combined with self-training, our approach underperforms with respect to the other policies. This result is expected, as our policy purposely explores the environment to find conflicting observations. Therefore, off-the-shelf predictions contain disagreeing pseudo-labels, which leads to poor results. Both SEAL and our perception are able to solve the inconsistencies. Nonetheless, we show that our perception strategy fully exploits the inconsistencies and outperforms SEAL perception by $0.70\%$.

In Table \ref{tab:results_policies}, we conduct an additional study by fine-tuning with ground truth labels, which can be interpreted as an \textit{upper-bound}. Not surprisingly, training with ground-truth labels outperforms pseudo-labels for every policy. \textit{Random goals} perform better than the other baselines ($+2.14\%$ compared to the second best baseline). Our intuition is that exploring by reaching random goals produces the highest variance in training data. If an oracle provides annotation, high variance is beneficial. 
Nevertheless, our approach scores the highest performance even when labels are provided, reaching mAP $48.01\%$.
\\
\subsection{Ablations}
\label{sec:ablations}
We study the impact of our instance-matching loss and consistent logits on performance and disagreement reward. We report additional ablations in Supplementary material.

\paragraph{\textbf{{Instance matching.}}}
We expect our instance-matching loss to benefit from an exploration policy that increases the number of views for an object. Table \ref{tab:abl} reports the results of applying instance-matching in conjunction with different policies. We notice that the object-detector benefits from the contrastive approach in conjunction with all the policies. Moreover, when backed by our policy, it shows the highest absolute improvement, with an increase in mAP@50 by $2.40\%$. 

\paragraph{\textbf{{Distillation loss.}}} We ablate the contribution of the aggregated soft consistent logits by comparing different values of $\alpha$, which regulates the weight of the distillation loss $\mathcal L_\text{im}$ and thus the training signal derived by injecting the consistent logits $\overline \lambda$ with a soft-distillation loss. All policies, except for SEAL and Semantic Curiosity, improve their performance if we introduce $\mathcal{L}_\text{distil}$, even when its weight is only $\alpha=0.1$. As discussed in Section \ref{sec:perception}, we expect the soft-distillation to improve the training when it does not exceed the cross-entropy loss for hard pseudo-labels. We find that weighting $\mathcal{L}_\text{distil}$ with $\alpha=0.7$ is the best option: \textit{our policy + our perception} outperforms the baselines and achieves the state-of-the-art result. On the other hand, $\mathcal{L}_\text{distil}$ lowers the mAP when backed by SEAL or Semantic Curiosity policies. This result is still not surprising. As discussed in Section \ref{sec:experiments}, policies with limited variance or training data cannot benefit from injecting the consistent logits through $\mathcal{L}_\text{distil}$.

\paragraph{\textbf{{Disagreement reward}}}
We conduct an ablation to compare the proposed disagreement scores with the following baselines:
\begin{itemize}
\item $s_\text{cos}$ is the average of the cosine distance between all the logits assigned to an object. Intuitively, we want the policy to maximize the distances between the logits assigned to each object. As the logits norm could be uninformative, we propose to adopt the cosine distance measure.

\item $s_\text{euc}$ is the average Euclidean distance between all the logits assigned to an object. As in $s_\text{cos}$, we measure disagreement as a distance; differently from  $s_\text{cos}$, $s_\text{euc}$ does take the norm of the logits into account.

\item $s_\text{count}$ is a disagreement score adapted from Semantic Curiosity \cite{chaplot2020semantic}. It is computed by counting the unique classes assigned to each object instance.

\end{itemize}

We compare the mAP of the object detector when we train the policy with various disagreement scores. We adopt \pname~as perception method. In our experiments, $s_\text{disag-entropy}$ performs the best, achieving mAP of $46.60\%$. $s_\text{cos}$ achieves $45.65\%$ mAP, $s_\text{euc}$ reaches $42.44\%$ mAP, while $s_\text{count}$ performs the worst with $40.58\%$ mAP. We highlight that our entropy-based disagreement is well grounded in the literature on epistemic uncertainty \cite{gal2016dropout,abdar2021review} and it is therefore well suited for estimating the uncertainty of the detector. $s_\text{cos}$ performs slightly worse than $s_\text{disag-entropy}$ but outperforms $s_\text{euc}$. Intuitively, $s_\text{cos}$ is more robust than $s_\text{euc}$ because the norm of the logits is uninformative for estimating disagreement between the predictions. On the contrary, $s_\text{count}$--adapted from \cite{chaplot2020semantic}--naively counts each object instance and fails to estimate uncertain cases for the object detector properly.

\subsection{Tests with a real humanoid robot}    
\label{sec:robot}

\begin{figure}[tb]
\centering
\begin{subfigure}{0.15\textwidth}
    \includegraphics[width=\textwidth]{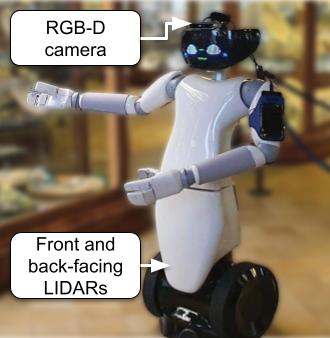}
    \caption{}

\end{subfigure}
\begin{subfigure}{0.15\textwidth}
    \includegraphics[width=\textwidth]{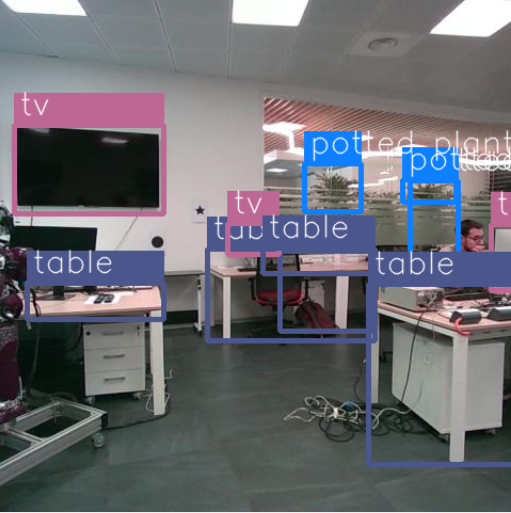}
    \caption{}

\end{subfigure}
\begin{subfigure}{0.15\textwidth}
    \includegraphics[width=\textwidth]{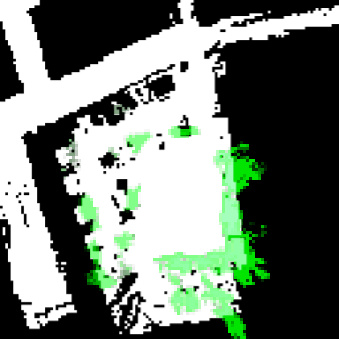}
    \caption{}
 
\end{subfigure}
\vspace{-10pt}
\caption{(a) Robot and sensor placement; (b) RGB image with detections superimposed; (c) map of the environment with disagreement superimposed in green.}
\label{fig:example_real}
\vspace{-.6cm}
\end{figure}

We finally experiment with the generalization of our approach in a real setting.
We deploy a policy that was trained only in the simulator on a wheeled humanoid robot and collect samples by exploring two different environments: an open-space area and an office-like environment with rooms and corridors.
The robot is R1 \cite{parmiggiani2017design} and is equipped with a Realsense D455 RGB-D camera mounted on its head at the height of 1.2m (Figure \ref{fig:example_real}). The robot is controlled via YARP \cite{yarp}, and the path planner used to follow goals and avoid obstacles is part of the ROS2 library \cite{ros}. %

We compare \aname with two baselines: random goals and human guidance. In the latter, an operator manually sends a series of goals to the robot, intending to see as many objects as possible from different views. 
In this setting, we fine-tune pre-trained MaskRCNN with \pname, self-training \cite{yalniz2019billion}, and SEAL perception \cite{chaplot2021seal}.
We test the object-detector on a set of 20 annotated images, of which 50\% in set and 50\% out of set.

Results are shown in Table \ref{tab:results_robot}. As a reference, off-the-shelf MaskRCNN achieves mAP@50 55.83.
\aname with \pname achieves an improvement of 9.97\% in mAP. We underline that this encouraging transfer ability is due to the policy being fairly agnostic to the input domain as well as the dynamics of the agent. In particular, the policy inputs sit at a higher level of abstraction than the agent's camera, being top-down grid maps; the policy outputs are also abstracted from the robot's dynamics, being goals in 2D space. 
On the other hand, manually collecting examples (human guidance) leads to worse results after fine-tuning. Humans struggle to solve tasks that are not clearly defined, e.g., collecting useful hard samples for fine-tuning.
\pname greatly outperforms the other perception methods, demonstrating its ability to improve the object-detector without relying on human annotations.

\begin{table}[t]
\caption{We deploy our policy on a humanoid robot and fine-tune MaskRCNN with our action-perception approach, and report the mAP. As reference, off-the-shelf detector reaches mAP@50 $55.83$.}
\centering
\begin{tabular}{l|lll}
\hline
\hline
Policy & Self-training & SEAL perception & \textbf{Disag. Reconc.} \\
\hline\hline
Random & 48.82 &54.49  &56.65 \\  %
Human guidance & 46.06 & 51.50 &  56.00\\
\hline
\baname & \textbf{53.47} & \textbf{56.85} & \textbf{65.62} \\ 
\end{tabular}
\label{tab:results_robot}
\vspace{-.6cm}
\end{table}

\section{Conclusion}
\label{sec:conclusion}
Our study introduced \aname, a novel exploration method, and \pname, a label-free fine-tuning approach for object detectors. Our method employs object detector uncertainty to derive a disagreement score, which, when maximized, effectively guides exploration and improves object detection in a real robotic test.
Our experiments demonstrate that \aname significantly outperforms both classic and learned exploration policies. By combining our uncertainty-based reward and disagreement map with the \pname module, we achieve a notable increase in mean Average Precision (mAP) of 46.60\%, a 6.27\% improvement over the baseline off-the-shelf object detector, and a 3.59\% advancement over state-of-the-art methods.
The experimental validation on a real robot is an important sanity check and further establishes the efficacy of our approach. We tested \pname with \aname on a real robotic platform without further tuning. In this scenario, our approach improved the object detector by 9.97\% without any human intervention.
We plan to release the implemented benchmark upon the acceptance of our paper. 

\textbf{Limitations and future works.}
In this work, we adopt Mask-RCNN as object detector for a fair comparison with previous approaches, as they adopt the same detector. Results can be improved by adopting other SOTA object detectors. Additionally, the current framework assumes a static environment, which may not accurately represent real-world scenarios where dynamic elements are present. Future work could focus on adapting our approach to more complex, changing environments, further enhancing its applicability and robustness in real-world applications.

\section*{Acknowledgements} This project has received funding from the European Union's Horizon research and innovation programme G.A. n. 101070227 (CONVINCE).
This work is supported by the project Future Artificial Intelligence Research (FAIR) – PNRR MUR Cod.
PE0000013 - CUP: E63C22001940006. This work is licensed under a Creative Commons Attribution 4.0
License. For more information, see https://creativecommons.org/licenses/by/4.0/, authors retained
Copyright of the material.

\bibliographystyle{splncs04}
\bibliography{main}

\clearpage
\maketitlesupplementary
\section{Additional details}

\paragraph{Implementation details on $\bm{\pi}$.} The policy network encoder is implemented in our experiments as a lightweight convolutional network, composed by 5 2D convolutional layers. The agent's orientation $e_t$ is fed through a linear embedding layer and concatenated to the convolutional feature. This is then followed by 2 linear layers. We train the policy by implementing an actor-critic agent with PPO \cite{schulman2017proximal}.

\paragraph{Training details} For the policy, we adopt the Adam optimizer \cite{kingma2017adam} with learning rate $2.5e^{-5}$, a discount factor $\gamma=0.99$, an entropy coefficient $0.001$, value loss coefficient $0.5$, and re-planning steps $N_\text{replanning}=40$, as in previous works \cite{chaplot2021seal,chaplot2020semantic}. We train all the policies for $250,000$ steps.
For the self-training stage, we adopt SGD optimizer with learning rate $\textit{lr}=1e^{-4}$, $\text{epochs} = 10$, $\text{weight-decay} = 1e^{-5}$, $\text{momentum} =0.9$ and $\text{batch size} = 16$. Intuitively, the soft-distillation loss $\mathcal L_\text{distil}$ should not dominate over cross-entropy, thus leading to incompatibilities between the two. We choose to set the soft-distillation weight $\alpha=0.7$ by empirically comparing the gradients of the two losses.
For the triplet-loss, we adopt the default $\text{margin}=0.3$.
We adopt train and test splits as in \cite{chaplot2021seal}. The code is available at \url{https://github.com/IIT-PAVIS/Look_Around_And_Learn}.

\section{Additional details on semantic voxel map creation}
\label{sec:voxelmapcreation}
\label{sec:voxelmapreprojection}
\begin{figure}[tbh]
\centering
\includegraphics[width=\linewidth]{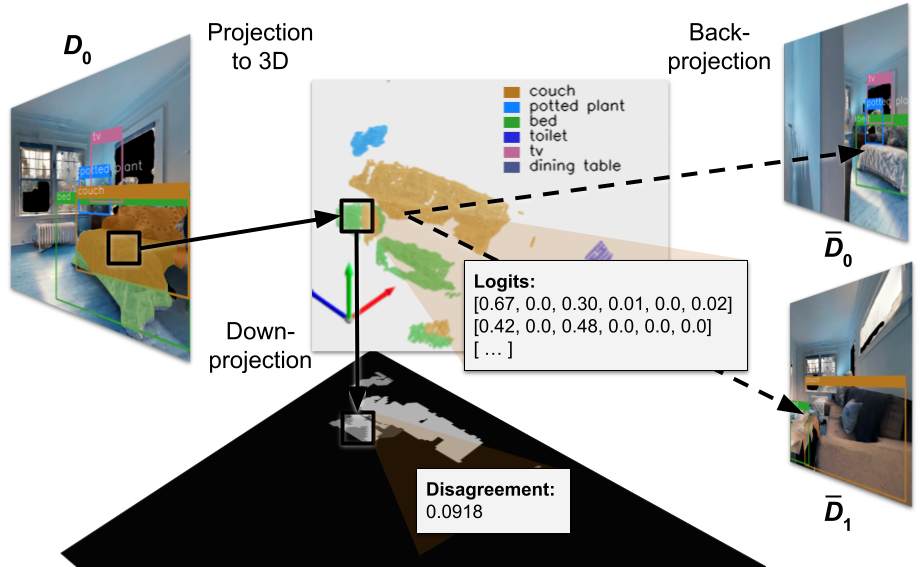}
\caption{Semantic voxel map creation and projection of detections onto 2D frames. First, we aggregate detections $D_0, \dots, D_N$ into semantic voxel-map. We solve the inconsistency of the voxel-map by assigning to each voxel the class with maximum score among the predictions of the voxel. Next, we project the semantic voxel-map back onto each observation, obtaining $\overline D_0, \dots, \overline D_N$. $\overline D_N$ is the consistent pseudo-label for observation $N$ and is obtained by reprojecting the voxel-map onto RGB-D frame $x_N$. Each pseudo-label $\overline D_i$ is associated to an object instance via the identifier $u_i$ and contains the consistent logits vector $\overline \lambda_{u_i}$.}
\label{fig:voxelmap-reprojection}
\end{figure}

Figure \ref{fig:voxelmap-reprojection} shows how detections are accumulated into a semantic voxel map, how we generate of the disagreement map from the semantic map, and finally how we reproject consistent pseudo-labels onto observations.
After each detection, instance segmentation masks $m_i$ at time $i$ are accumulated into a dense semantic point cloud by lifting 2D points to 3D using the agent's global pose and camera intrinsics. We assign to each 3D point its corresponding predicted class and score logits vector as predicted by the detector.
The point cloud is then subsampled via voxelization using the Pyntcloud library into a voxel map with leaf size $0.05m$. At the same time, we solve inconsistencies by assigning to each voxel the class label $\overline{y}$ with the maximum score among all predictions for that voxel.
We adopt a 3D connected component algorithm to assign connected voxels of the same predicted object class $\overline{y}$ to the same 3D object instance, where each 3D object instance is identified with a unique object identifier $u$. Object ids point to all logits $\lambda$ corresponding to the detections accumulated for that object, which are stored separately.
The voxel map is then condensed into the disagreement map $H$ as follows. We compute the disagreement scores $s_\text{disag-entropy}(u)$ as in Equation {\color{red}1} for all objects $u$. 
Next, we sample points $(i,j)$ on a regular grid along the 2D coordinate plane. For each $(i, j)$, we sum the disagreement scores of the object lying inside the cell and assign the total score to $H_i(i, j)$.
Finally, during fine-tuning, the objects (with their consistent labels) are projected back into all collected observations via raytracing.

\section{Qualitative results}
\label{sec:qualitativegibson}
\begin{figure}[t] 
\centering
\begin{subfigure}[b]{.32\linewidth}
\includegraphics[width=\linewidth]{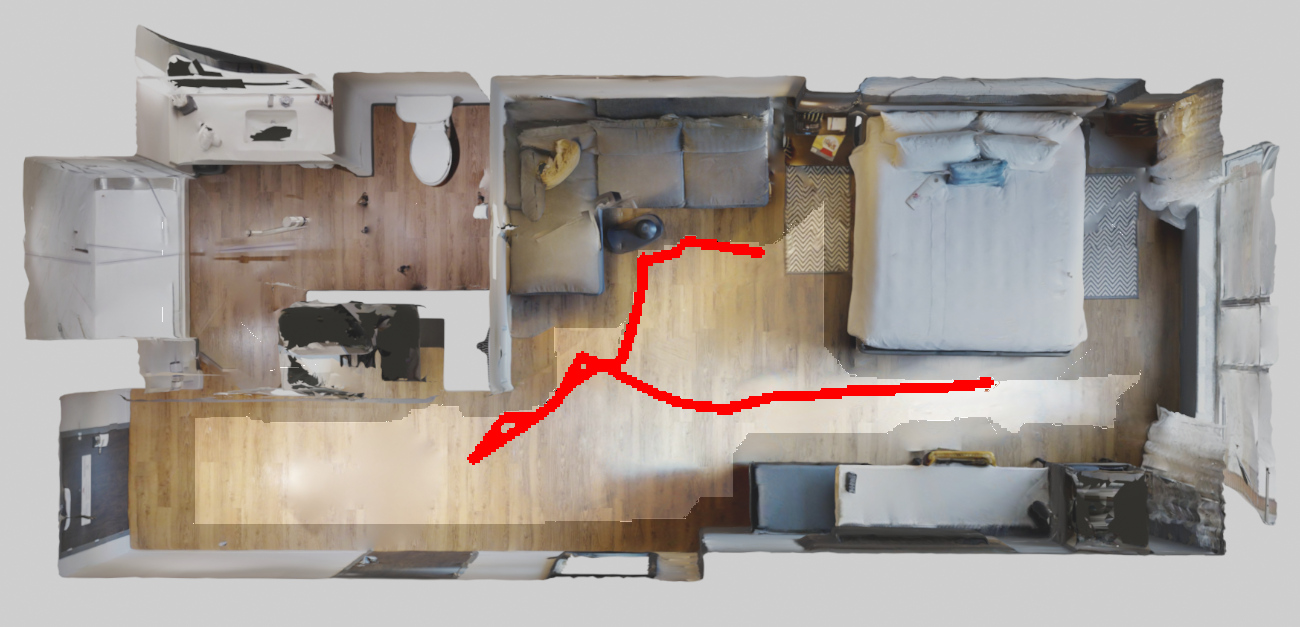}
\caption{Random goals}
\end{subfigure} 
\begin{subfigure}[b]{.32\linewidth}
\includegraphics[width=\textwidth]{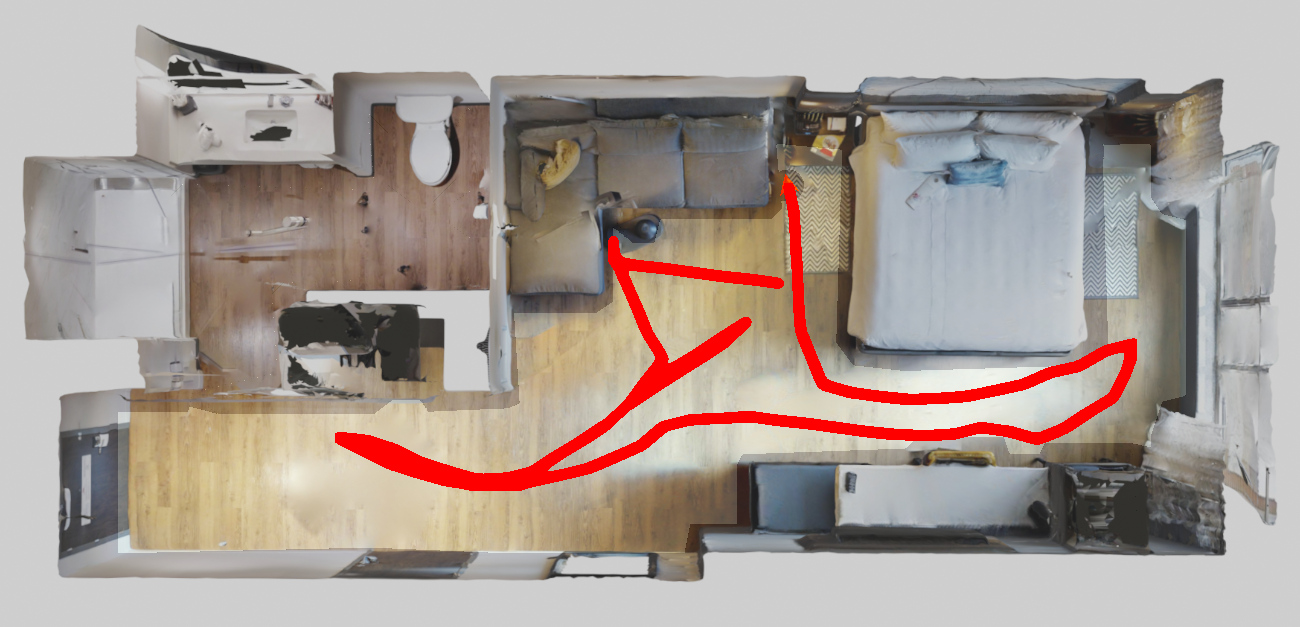}
\caption{Frontier \cite{yamauchi1998frontier}}
\end{subfigure} 
\begin{subfigure}[b]{.32\linewidth} 
\includegraphics[width=\textwidth]{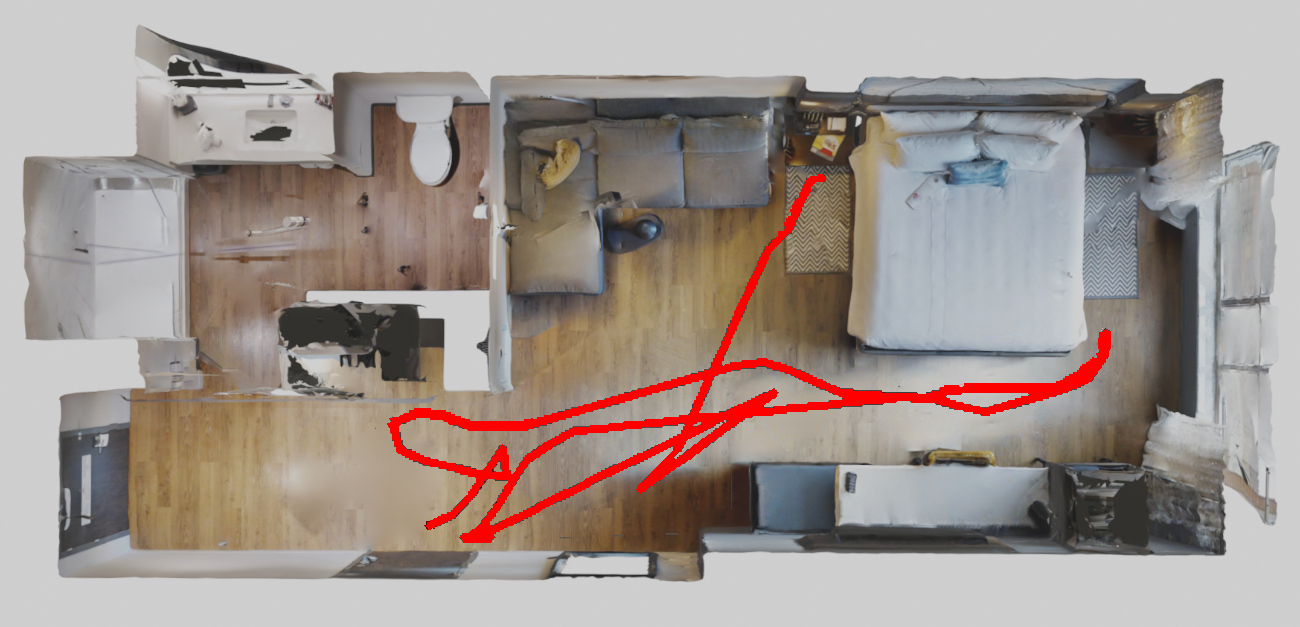}
\caption{NeuralSLAM \cite{chaplot2020learning}}
\end{subfigure}
\begin{subfigure}[b]{.32\linewidth}
\includegraphics[width=\textwidth]{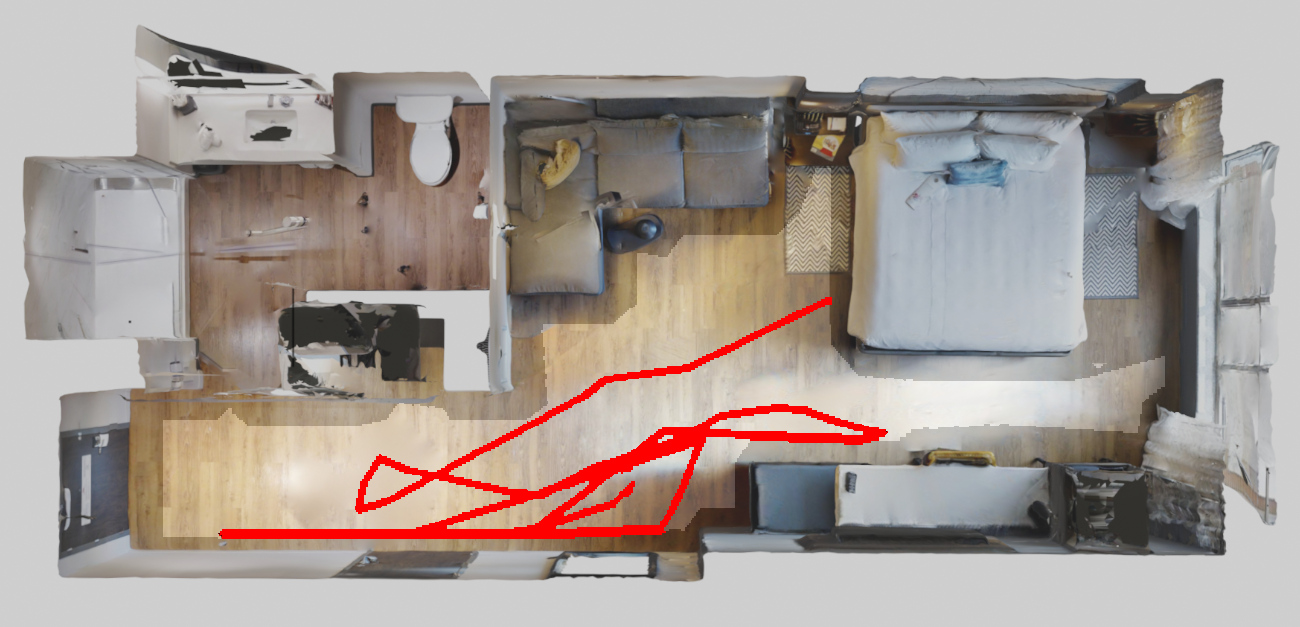}
\caption{SEAL \cite{chaplot2021seal}}
\end{subfigure} 
\begin{subfigure}[b]{.32\linewidth}
\includegraphics[width=\textwidth]{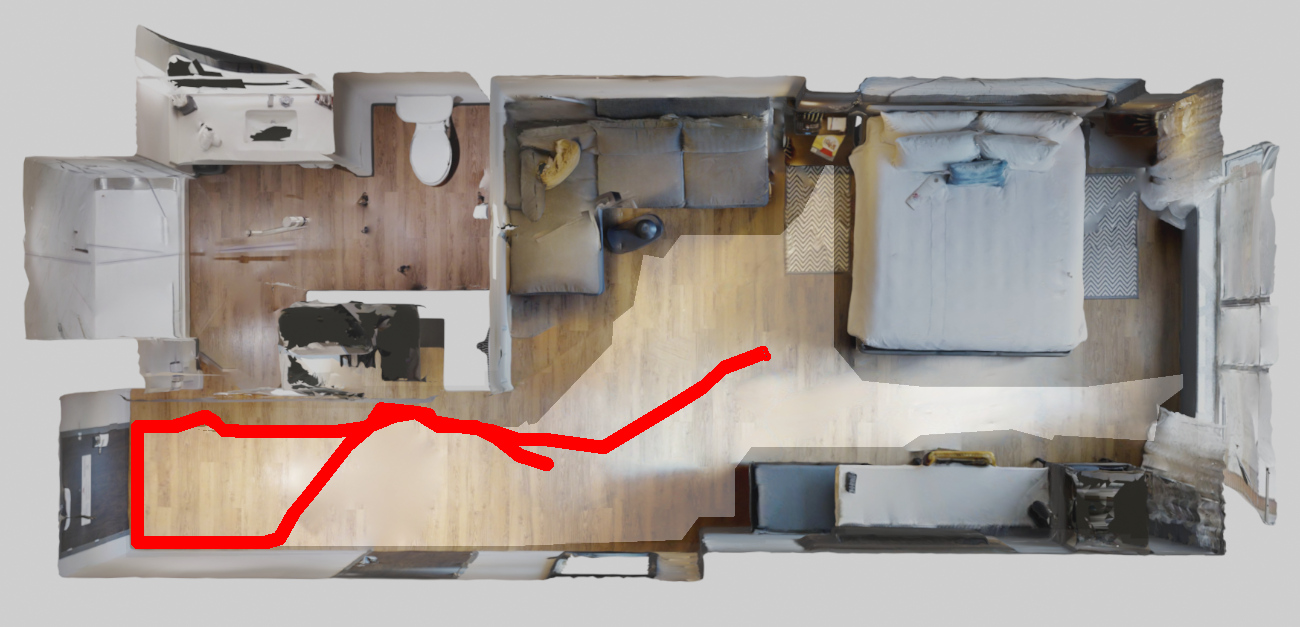}
\caption{Sem. Cur. \cite{chaplot2020semantic}}
\end{subfigure} 
\begin{subfigure}[b]{.32\linewidth} 
\includegraphics[width=\textwidth]{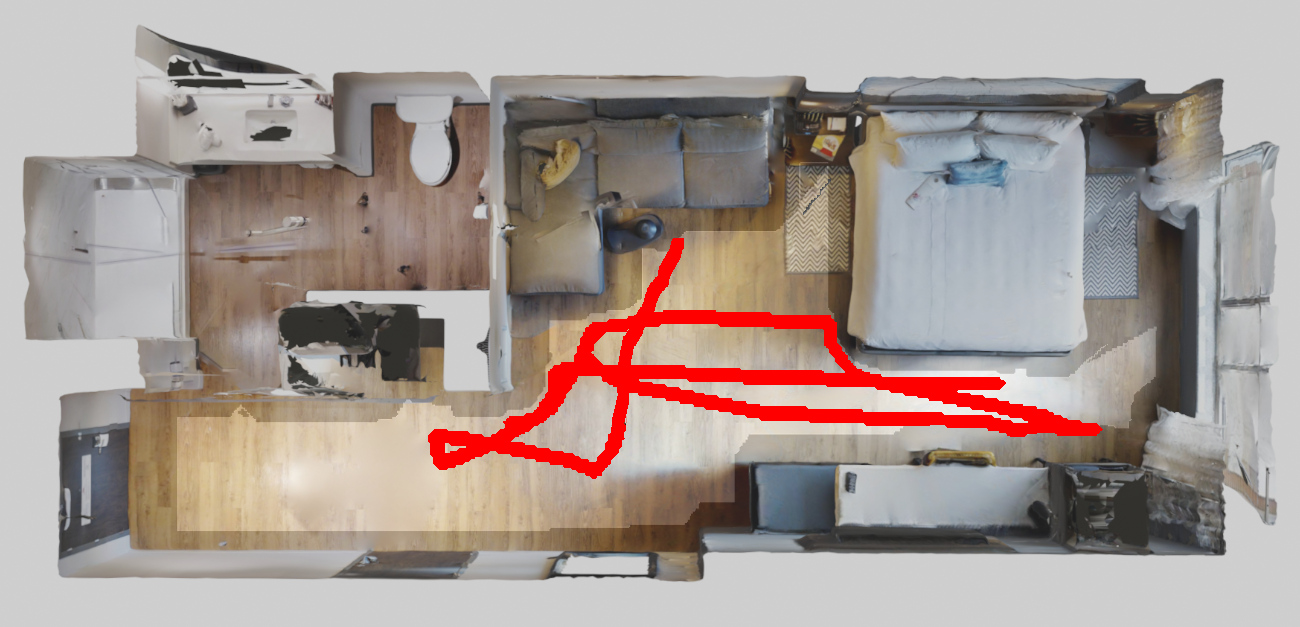}
\caption{Ours}
\end{subfigure} 
\hfill
\caption{We sample a trajectory of 300 steps for a scene that is not part of the training scenes (Pablo \cite{xiazamirhe2018gibsonenv}). The agent trajectory is shown as a red line. The floor area covered by the agent is highlighted. All policies are capable of exploring a good portion of the environment.
}
\label{fig:policies}
\end{figure}

We investigate qualitatively the behaviors of different policies. Figure \ref{fig:policies} shows an example trajectory from each policy on an unseen scene (Pablo). Frontier and NeuralSLAM try to maximize explored space directly, while SEAL's reward is proportional to the number of detected objects. On the other hand, our policy does not directly reward exploration.
Although all policies cover most of the walkable area of the scene, we show in the experimental section that maximizing exploration is sub-optimal for the task of increasing object-detector performance.

\begin{figure*}[tb] \centering
\includegraphics[width=0.8\linewidth]{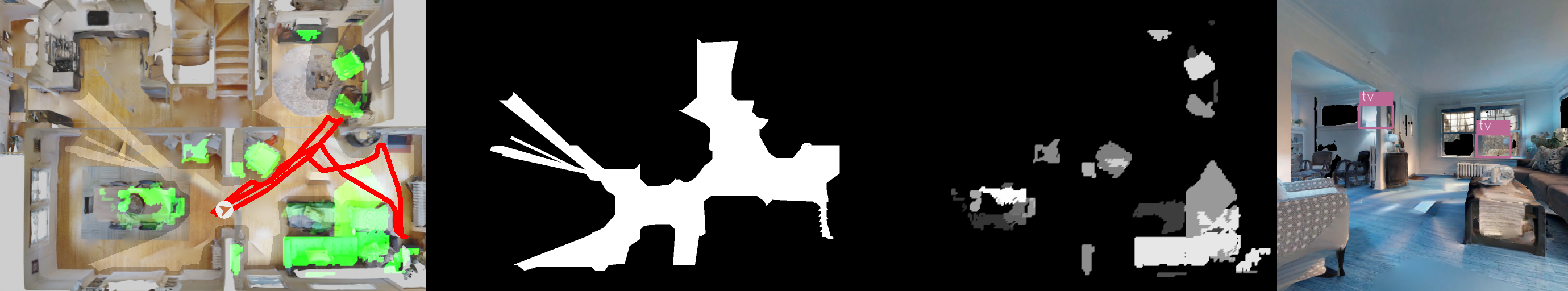}
\caption{Our policy successfully explores the environment and builds an extensive disagreement map. The agent visits uncertain objects multiple times, accumulating samples.}
\label{fig:success}
\end{figure*}

\begin{figure*}[tbh] \centering
\begin{subfigure}[b]{\linewidth} \centering
\includegraphics[width=0.8\linewidth]{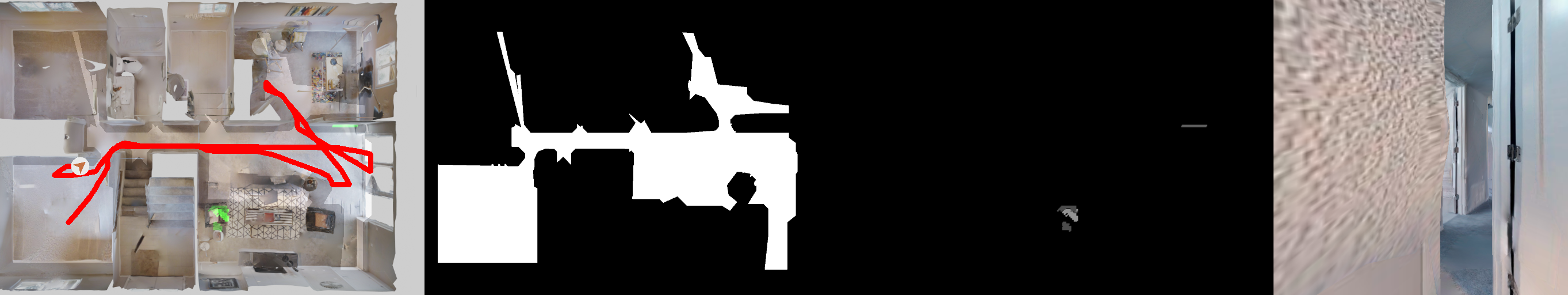}
\caption{}
\label{fig:failurea}
\includegraphics[width=0.8\linewidth]{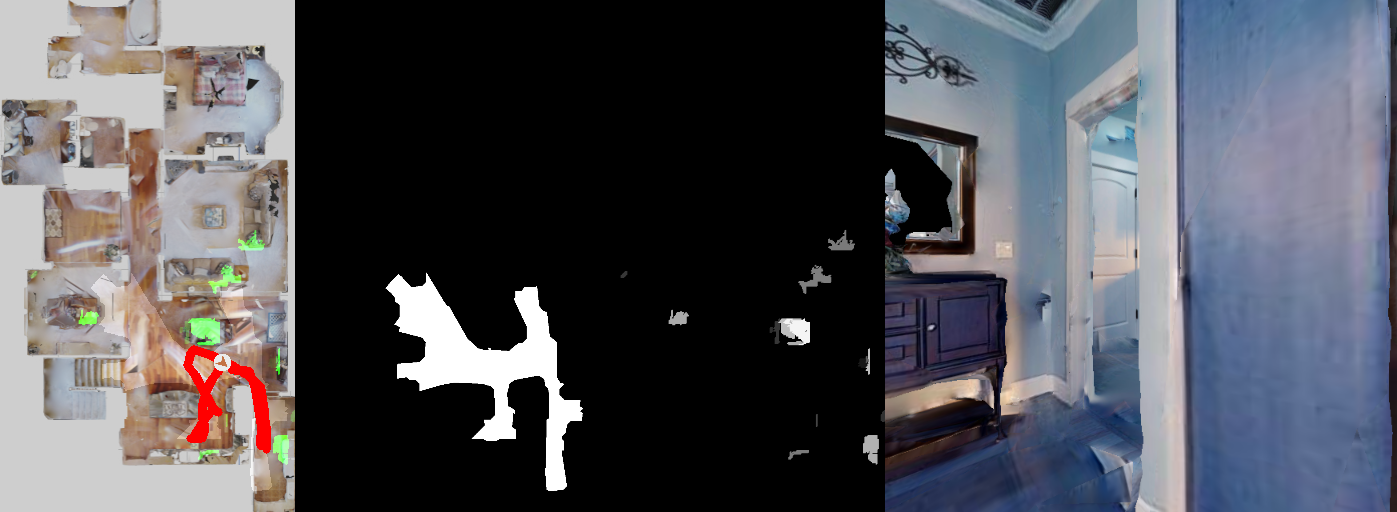}
\caption{}
\label{fig:failureb}
\includegraphics[width=0.8\linewidth]{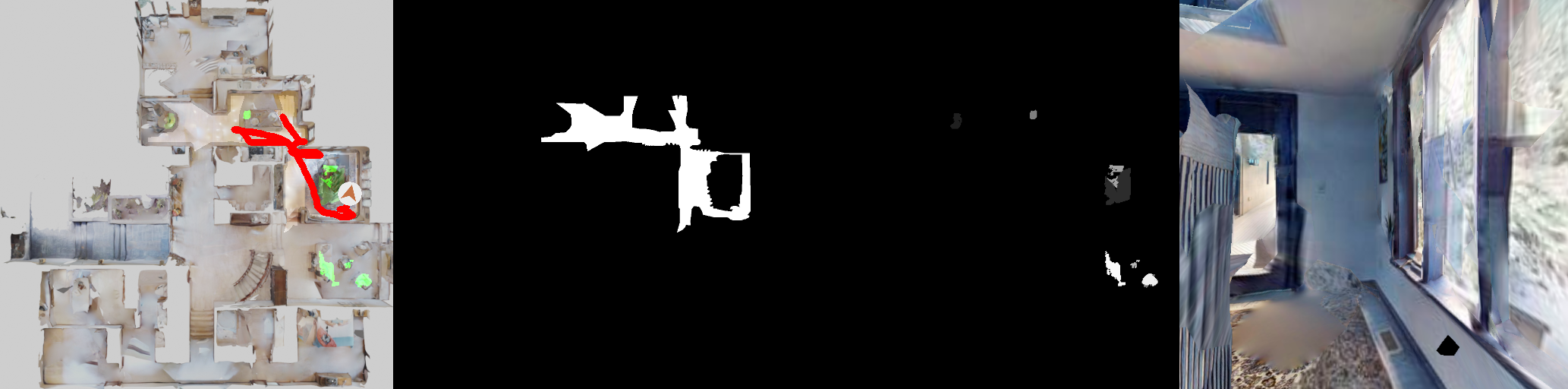}
\caption{}
\label{fig:failurec}
\includegraphics[width=0.8\linewidth]{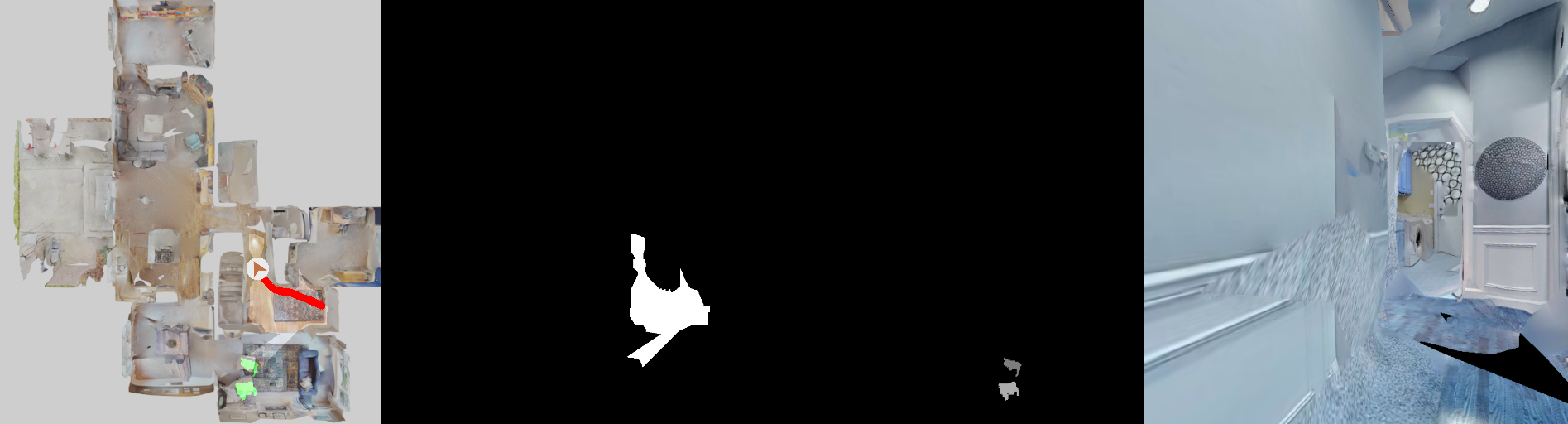}
\caption{}
\label{fig:failured}
\end{subfigure} 

\caption[]{Failure cases: \textit{(a)} Few objects are detected and the disagreement map does not grow; \textit{(b)},\textit{(c)} the agent gets stuck in a small area and keeps visiting the same objects; \textit{(d)} corrupted mesh of the scene prevents the agent from exploring further.}
\label{fig:failure}
\end{figure*}

We show some successful and failed examples of our policy in action. Figure \ref{fig:success} shows a successful data collection episode. The policy explores the scene while building a disagreement map containing several objects. 
On the other hand, Figure \ref{fig:failure} shows some typical failure cases for our policy. 
In Figure \ref{fig:failurea}, when few objects of the class list are present or detected in the map, the resulting disagreement map is mostly empty.
A typical failure mode occurs when the agent remains stuck in a small area and keeps visiting the same objects throughout the episode (Figure \ref{fig:failureb}, \ref{fig:failurec}).
Finally, in some rare instances the environment mesh presents glitches (Figure \ref{fig:failured}) which can prevent the agent from exploring in a certain direction, with the agent being unable to plan an alternative route.

\end{document}